\documentclass[sigconf]{acmart}
\AtBeginDocument{%
  }

\definecolor{Gray}{HTML}{C8E8FA}
\usepackage{xcolor, colortbl}
\usepackage{tablefootnote}
\usepackage[utf8]{inputenc}
\usepackage{lipsum} 
\usepackage[utf8]{inputenc}
\usepackage{graphicx}  
\usepackage{multirow}  

\usepackage{framed}

\newenvironment{chatgptbox}{%
  \def\FrameCommand##1{%
    \setlength{\fboxsep}{5pt}
    \colorbox{green!5}{##1}
  }%
  \MakeFramed{\advance\hsize-\width \FrameRestore}%
  \ttfamily\small
}{\endMakeFramed}

\copyrightyear{2025}
\acmYear{2025}
\setcopyright{cc}
\setcctype{by}
\acmConference[K-CAP '25]{Knowledge Capture Conference 2025}{December 10--12, 2025}{Dayton, OH, USA}
\acmBooktitle{Knowledge Capture Conference 2025 (K-CAP '25), December 10--12, 2025, Dayton, OH, USA}
\acmDOI{10.1145/3731443.3771357}
\acmISBN{979-8-4007-1867-0/2025/12}
\begin{document}

\title{Minimizing Hyperbolic Embedding Distortion with LLM-Guided Hierarchy Restructuring}

\author{Melika Ayoughi}
\email{m.ayoughi@uva.nl}
\orcid{0000-0002-6847-5999}
\affiliation{%
  \institution{University of Amsterdam}
  \country{The Netherlands}
}

\author{Pascal Mettes}
\email{p.s.m.mettes@uva.nl}
\orcid{0000-0001-9275-5942}
\affiliation{%
  \institution{University of Amsterdam}
  \country{The Netherlands}
}

\author{Paul Groth}
\email{p.t.groth@uva.nl}
\orcid{0000-0003-0183-6910}
\affiliation{%
  \institution{University of Amsterdam}
  \country{The Netherlands}
}

\renewcommand{\shortauthors}{Ayoughi et al.}

\begin{abstract}
Hyperbolic geometry is an effective geometry for embedding hierarchical data structures. Hyperbolic learning has therefore become increasingly prominent in machine learning applications where data is hierarchically organized or governed by hierarchical semantics, ranging from recommendation systems to computer vision. The quality of hyperbolic embeddings is tightly coupled to the structure of the input hierarchy, which is often derived from knowledge graphs or ontologies. Recent work has uncovered that for an optimal hyperbolic embedding, a high branching factor and single inheritance are key, while embedding algorithms are robust to imbalance and hierarchy size. To assist knowledge engineers in reorganizing hierarchical knowledge, this paper investigates whether Large Language Models (LLMs) have the ability to automatically restructure hierarchies to meet these criteria. We propose a prompt-based approach to transform existing hierarchies using LLMs, guided by known desiderata for hyperbolic embeddings. Experiments on 16 diverse hierarchies show that LLM-restructured hierarchies consistently yield higher-quality hyperbolic embeddings across several standard embedding quality metrics. Moreover, we show how LLM-guided hierarchy restructuring enables explainable reorganizations, providing justifications to knowledge engineers.
\end{abstract}




\maketitle
\section{Introduction}

Hyperbolic learning is increasingly common in machine learning applications ranging from computer vision \cite{li2024isolated,mettes2024hyperbolic} to recommendation systems \cite{tu2024hyperbolic,yang2022hicf} and knowledge graph completion \cite{kolyvakis2020hyperbolic,li2024hyperbolic}. The quality of hyperbolic embeddings is tightly coupled to the input hierarchy, which often stems from knowledge graphs and ontologies \cite{Wang2013,ayoughi2025designing,nickel2017poincare}. Hence, knowledge engineers can have a large impact on downstream performance through their choices in how they capture and structure their ontologies. 

To help guide knowledge engineers, Ayoughi et al. \citep{ayoughi2025designing} proposed a series of recommendations for them in their design of hierarchies based on an extensive set of experiments. This paper builds on these recommendations by investigating whether Large Language Models (LLMs) can provide effective assistance to restructure hierarchies to improve their embedding quality. LLMs have a number of potential benefits in terms of assisting knowledge engineers. First, LLMs can be used to flexibly apply natural language recommendations to a variety of ontologies from different domains, taking advantage of pretrained world knowledge. Second, LLMs can be used with multiple different formats, making it easy to use different knowledge graph representations (e.g., RDF, property graphs, textual formats) \cite{meyer2025llm}. Third, LLMs can produce explanations as to why changes are recommended, allowing knowledge engineers to either adopt the changes directly or make trade-offs with other representation goals \cite{lubos2024llm}. Our investigation makes the following key contributions:
\begin{enumerate}
    \item A prompt-based approach using LLMs that transforms existing hierarchies according to the recommendations of \citep{ayoughi2025designing} while generating associated explanations. 
    \item A comparison of the quality of embeddings produced from 16 different hierarchies before and after restructuring using the aforementioned approach. The hierarchies stem from domains ranging from robotics to biodiversity. 
    \item A characterization of the explanations that the approach produces.
\end{enumerate}

Our results show that across domains, our LLM-based restructuring approach produces higher-quality embeddings in terms of both average and worst-case distortion. In addition, the approach provides clear explanations for the reconstruction to the knowledge engineer. Our code and data is openly available here \footnote{\url{https://github.com/Melika-Ayoughi/LLM-guided-Hierarchy-Restructuring/}}.
\section{Related Work}
\subsection{Prompt-based knowledge engineering}
Our work is in line with the recent trend on ontology and knowledge engineering using Large Language Models (LLMs) \cite{lippolis2025ontology, allen2023knowledge, li2025large, shimizu2025accelerating,hertling2023olala}. LLMs have proven to be flexible, have good performance, and support knowledge engineers in a range of tasks \cite{val2025ontogenix,zhang2024ontochat,koutsiana2024knowledge}. In particular, the use of prompt engineering approaches has become popular for ontology generation \cite{saeedizade2024navigating,lippolis2024ontogenia, babaei2023llms4ol,soares2025exploring} and ontology completion \cite{kollapally2025ontology,xia2023find,xu2025compress}. Our work is different in that we focus on restructuring existing hierarchies and measuring subsequent performance on hierarchical machine learning.

\subsection{Ontology design}
Hierarchical structures—particularly taxonomic backbones, whether formally defined or informally constructed—are fundamental to the design of ontologies and knowledge graphs \cite{hofer_construction_2024,kendall_ontology_2019,ontologies_2003}. They provide a means to decompose complex domains into modular components \cite{shimizu_modular_2023}, facilitating various forms of reasoning such as subsumption. In the context of ontology induction and knowledge graph construction, building high-quality hierarchies is a central consideration \cite{hierarchical_2022,weikum_machine_2021}. Their evaluation commonly relies on expert validation, alignment with gold-standard ontologies, or adherence to established evaluation criteria \cite{mcdaniel_evaluating_2020}. Given that modifications to hierarchical structures can propagate and substantially affect downstream applications \cite{pernisch_beware_2021}, ontology design methodologies offer guidance for structuring hierarchies to reflect domain-specific constraints and support correct reasoning outcomes \cite{kendall_ontology_2019,poveda-villalon_lot:_2022}.

Ontology evaluation studies frequently consider the hierarchical structure \cite{gmez-prez_evaluation_2001,mcdaniel_evaluating_2020}. These studies examine whether the hierarchy correctly partitions instances, avoids cycles of specialization and generalization, and supports semantically accurate instance assertions \cite{gmez-prez_evaluation_2001}. Other evaluation approaches adopt principled criteria and metrics based on formal notions (e.g., unity) \cite{beydoun_how_2011,guarino_evaluating_2002,guizzardi_types_2021} to assess the quality of a hierarchy. Common criteria include structural complexity (e.g., number of classes, hierarchy depth, number of top-level classes) and conciseness (e.g., absence of cycles, presence of instance-less classes) \cite{10.1145/3627673.3679156}. Recent work has also explored the impact of hierarchical design choices on downstream hyperbolic machine learning representations and has proposed design recommendations accordingly \cite{ayoughi2025designing}. Our work complements these recommendations, metrics, and evaluation approaches by providing ontology engineers with an LLM-guided assistant that can translate these guidelines and apply them to existing hierarchies, thereby potentially supporting more effective integration of hierarchical ontologies in hyperbolic deep learning.

\subsection{Hyperbolic embeddings and learning}
"Hyperbolic space can be thought of as a continuous analogue to discrete trees" \cite{nickel2018learning}, owing to their shared nature of exponential growth. Hyperbolic embeddings are the focus of this study because they demonstrate superior performance in representing hierarchical data structures compared to Euclidean methods. Hyperbolic embeddings can be obtained through gradient-based or construction-based approaches. In early work, Sarkar \cite{sarkar2011low} introduced 2D construction-based Delaunay tree embeddings in hyperbolic space, demonstrating the potential of hyperbolic geometry to achieve tree embeddings with arbitrarily low distortion.
Nickel and Kiela \cite{nickel2017poincare} addressed this limitation by proposing a contrastive approach that supports embedding optimization in any dimensionality, significantly outperforming Euclidean embeddings on trees. This line of work has been extended through explicit entailment losses \cite{ganea2018hyperbolic,yu2023shadow}, using different hyperbolic models \cite{nickel2018learning}, separation \cite{long2020searching} or distortion objectives \cite{yu2022skin}.
The current state-of-the-art methods extend Sarkar's constructive approach to higher-dimensional embeddings by positioning points recursively with pre-defined or uniform objectives \cite{sala2018representation,van2025low}. Ayoughi et al. \cite{ayoughi2025designing} showed that such constructive approaches consistently outperform gradient-based methods. Hence, in this work, we will use constructive hyperbolic embedding algorithms throughout our experiments.

In light of the strong performance of hyperbolic representation learning, numerous studies have integrated hyperbolic embeddings into neural networks, enabling deep learning to incorporate hierarchical knowledge. Hyperbolic learning have been shown to improve recognition across various domains, including image and video classification \cite{li2024isolated,liu2020hyperbolic,ayoughi2025continual}, word embeddings \cite{dhingra2018embedding,le2019inferring}, recommender systems \cite{tu2024hyperbolic,yang2022hicf}, audio understanding \cite{hong2023hyperbolic}, biological domains \cite{klimovskaia2020poincare}, networks and graphs \cite{yang2023hyperbolic,zhou2023hyperbolic}, and image-text settings \cite{poppi2025hyperbolic, desai2023hyperbolic,ibrahimi2024intriguing,pal2024compositional}. Beyond classification, hyperbolic representations facilitate hierarchical recognition \cite{dhall2020hierarchical,ghadimi2022hyperbolic}, learning from limited samples \cite{hong2023curved,ma2022adaptive,zhang2022hyperbolic,atigh2025simzsl}, interpretability \cite{gulshad2023hierarchical}, robustness \cite{li2024hyperbolic}, and other tasks. For a comprehensive overview of advances in hyperbolic learning, we refer to recent surveys \cite{mettes2024hyperbolic,peng2021hyperbolic}. In this work, we challenge the assumption that the hierarchical information is fixed \emph{a priori} and improve the hyperbolic embedding quality by restructuring the hierarchy.

\section{Approach: LLM-Guided hierarchy restructuring}
\begin{figure*}
  \centering \includegraphics[width=\textwidth, trim={0cm 11cm 0cm 0cm}, clip]{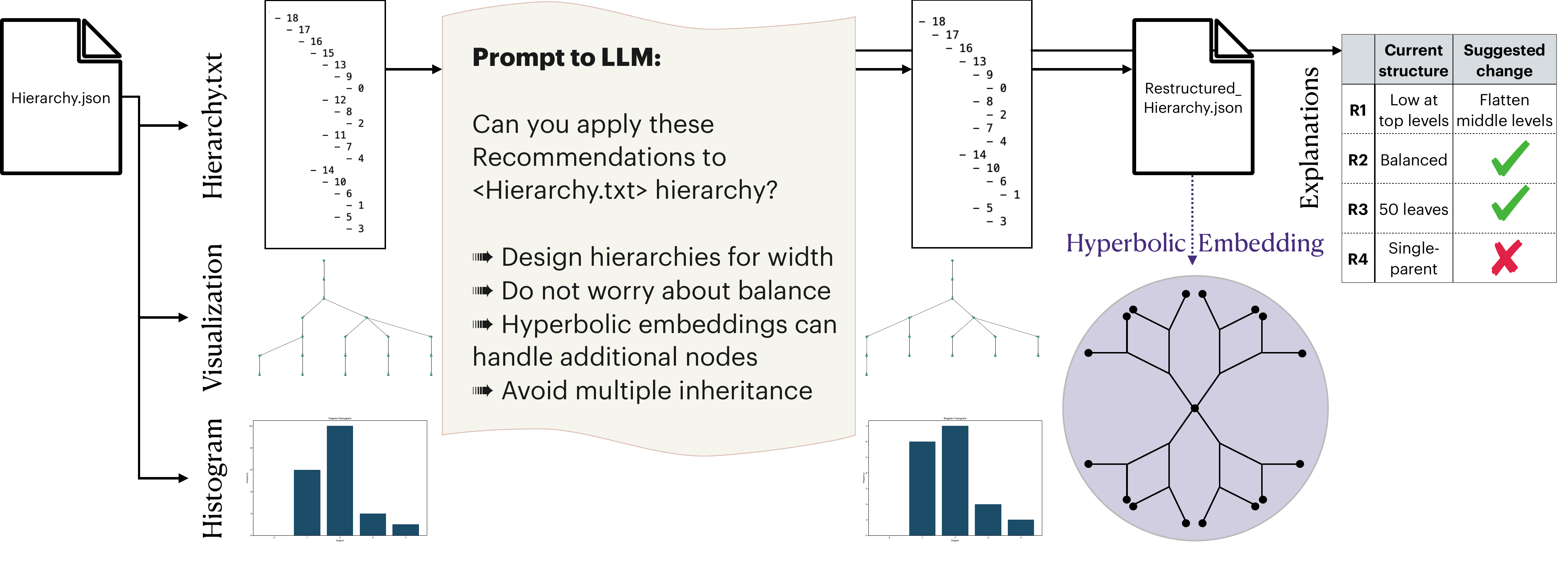}
  \caption{Prompt-based LLM-Guided hierarchy restructuring approach}
  \Description{}
  \label{fig:method}
\end{figure*}

Our approach, outlined in Figure \ref{fig:method}, is described below.

\paragraph{\textbf{Hierarchy reformatting}} Ontologies and knowledge hierarchies are represented in a wide range of formats. These formats can be challenging for language models (LLMs) due to (i) input token length constraints, and (ii) their inability to interpret such representations as structured trees. To overcome these limitations, we transform the original hierarchy into a compact, interpretable textual representation.

We generate the textual hierarchy using a pre-order depth-first traversal (DFS) that preserves the tree’s structure. This representation encodes each node’s position relative to its parent and siblings, effectively capturing the hierarchy’s shape, balance, depth, and branching factor. Alongside, we extract and store tree properties, including node count, edge count, leaf node count, branching factor histogram, and tree visualizations for both the original and transformed hierarchies.

\begin{figure}[t]
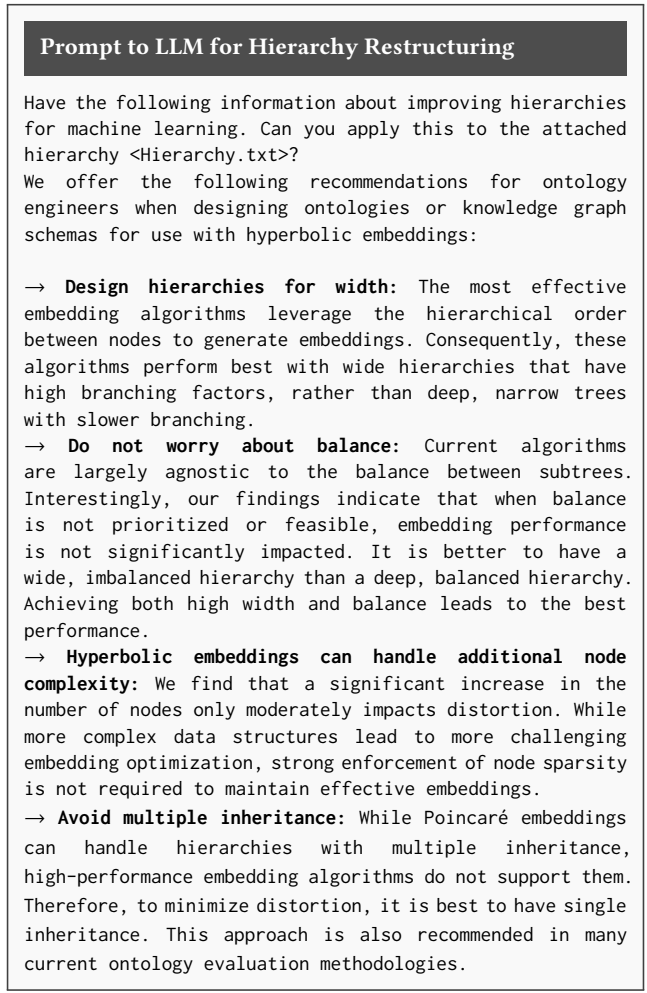

\centering

\setlength{\fboxsep}{6pt}   
\setlength{\fboxrule}{0.6pt}

\begin{minipage}{\linewidth}
\fcolorbox{black!75}{gray!5}{%
  \parbox{\dimexpr\linewidth-2\fboxsep-2\fboxrule\relax}{%
    {\setlength{\fboxsep}{6pt}
     \colorbox{black!70}{%
       \parbox{\dimexpr\linewidth-2\fboxsep\relax}{%
         \bfseries\color{white} Prompt to LLM for Hierarchy Restructuring
       }%
     }%
    }%

    \vspace{6pt}

    {\ttfamily\small
    Have the following information about improving hierarchies for machine learning. Can you apply this to the attached hierarchy <Hierarchy.txt>? 

We offer the following recommendations for ontology engineers when designing ontologies or knowledge graph schemas for use with hyperbolic embeddings:
\\\\
$\rightarrow$  \textbf{Design hierarchies for width:} The most effective embedding algorithms leverage the hierarchical order between nodes to generate embeddings. Consequently, these algorithms perform best with wide hierarchies that have high branching factors, rather than deep, narrow trees with slower branching.

$\rightarrow$ \textbf{Do not worry about balance:} Current algorithms are largely agnostic to the balance between subtrees. Interestingly, our findings indicate that when balance is not prioritized or feasible, embedding performance is not significantly impacted. It is better to have a wide, imbalanced hierarchy than a deep, balanced hierarchy. Achieving both high width and balance leads to the best performance.

$\rightarrow$ \textbf{Hyperbolic embeddings can handle additional node complexity:} We find that a significant increase in the number of nodes only moderately impacts distortion. While more complex data structures lead to more challenging embedding optimization, strong enforcement of node sparsity is not required to maintain effective embeddings.

$\rightarrow$ \textbf{Avoid multiple inheritance:} While Poincaré embeddings can handle hierarchies with multiple inheritance, high-performance embedding algorithms do not support them. Therefore, to minimize distortion, it is best to have single inheritance. This approach is also recommended in many current ontology evaluation methodologies.
    }
  }%
}
\end{minipage}

\caption{The prompt used for restructuring the hierarchy.}
\label{fig:prompt}
\end{figure}

\paragraph{\textbf{Prompting}}
To optimize the hierarchy for hyperbolic embeddings, we build upon the recommendations by Ayoughi et al.~\cite{ayoughi2025designing} that empirically identify structural properties favorable to low-distortion hyperbolic embeddings, including favoring wide, single-inheritance hierarchies, de-prioritizing the hierarchy size and imbalance. We prompt the LLM with the transformed textual hierarchy and instruct it to restructure the hierarchy to optimize for hyperbolic embedding quality, incorporating these recommendations. The prompt, shown in Figure \ref{fig:prompt}, includes the structural guidelines and the hierarchy text.

\paragraph{\textbf{Post prompt validation}} The LLM-generated hierarchy is subsequently parsed and analyzed to ensure four key criteria are met:

\begin{enumerate}
    \item The restructured hierarchy must differ structurally from the input hierarchy.
    \item All original leaf nodes must be retained, as these typically represent classification targets in downstream tasks.
    \item The nodes/edges and the hierarchy structure have not been hallucinated
    \item The format of the output is the same as the input hierarchy
\end{enumerate}

If any criterion is not satisfied, a follow-up prompt is issued to ensure that the criterion is met. If this does not work, the prompting process is repeated from the top. During the prompting process, the LLM is asked to produce an explanation. Once all criteria are fulfilled, the textual representation of the modified hierarchy is parsed into a graph dictionary format for further processing. The restructured hierarchy can then be used as input for various hyperbolic embedding methods, as we discuss next. 

\paragraph{\textbf{Hyperbolic embedding}}

For the purposes of embedding, we are given a tree-like data structure $T = (V, E)$, containing a set of nodes $V$ and a set of edges $E$, with each edge $e \in E$ connecting two vertices. Algorithms strive to obtain a continuous analogue of $T$ by embedding each node $v \in V$ in an embedding space, such that the distance between two nodes corresponds one-to-one to the shortest path between the nodes in the tree, as given by the number of edges between them. Let $\phi : V \mapsto \mathbb{D}^n$ denote the embedding function that takes nodes as input and outputs their embedding in an $n$-dimensional hyperbolic space $\mathbb{D}^n$.
Following \cite{ganea2018hyperbolic,nickel2017poincare,sala2018representation}, we operate in the Poincaré ball model of hyperbolic space for the embeddings. For an $n$-dimensional space, let $(\mathbb{D}^n, \mathfrak{g}^n)$ denote the Riemannian manifold of the Poincaré ball model, given as \(\mathbb{D}^n = \big\{ \mathbf{x} \in \mathbb{R}^n : ||\mathbf{x}||^2 < 1 \big\}\), \(\mathfrak{g}^n = \lambda_\mathbf{x} I_n\), where \(\lambda_\mathbf{x} = \frac{2}{1 - ||x||^2}\). A key operator for hyperbolic embedding algorithms is the distance between two vectors in hyperbolic space. Here, it denotes the distance between the embeddings of two nodes. For nodes $\mathbf{v}_1, \mathbf{v}_2 \in \mathbb{D}^n$, the distance is given as \(d_{\mathbb{D}} (\mathbf{v}_1, \mathbf{v}_2) = 2 \tanh^{-1} \big( ||-\mathbf{v}_1 \oplus \mathbf{v}_2|| \big),\) where $\oplus$ denotes the Möbius addition, defined as:
\begin{equation}\label{eq:mob_add}
    \mathbf{v}_1 \oplus \mathbf{v}_2 = \frac{(1 + 2 \langle \mathbf{v}_1, \mathbf{v}_2 \rangle + ||\mathbf{v}_2||^2) \mathbf{v}_1 + (1 - ||\mathbf{v}_1||^2) \mathbf{v}_2}{1 + 2 \langle \mathbf{v}_1, \mathbf{v}_2 \rangle + ||\mathbf{v}_1||^2 ||\mathbf{v}_2||^2}.
\end{equation}

This manifold and distance functions are used both to generate embeddings and in their evaluation, as discussed later. Our focus is on state-of-the-art construction-based methods such as Hadamard \cite{sala2018representation} and HS-DTE \cite{van2025low}, since they produce high-quality embeddings that preserve nearly all of the original tree structure \cite{ayoughi2025designing}. In contrast, gradient-based Poincaré Embeddings \cite{nickel2017poincare}, although applicable to any graph, lead to $10$ times worse embedding distortions in hierarchies\cite{ayoughi2025designing}. In short, both hyperbolic embeddings used in this paper build a continuous hyperbolic embedding of a hierarchy in a recursive manner. The root is placed at the origin, and its children are distributed around the root. Then, these children become the parent node, with their children distributed accordingly. The main difference between the methods lies in how the children are positioned. Sala et al. \cite{sala2018representation} use a fixed Hadamard code to place children at the vertices of a hypercube inscribed within the hypersphere. Van Spengler et al. \cite{van2025low} learn an approximate uniform distribution of children on the hypersphere around the parent through gradient descent.
\section{Experimental setup}
\paragraph{\textbf{Hierarchies}}
We apply the LLM-guided hierarchy restructuring pipeline to 16 diverse sets of hierarchies that encompass diverse structural properties. The number of nodes, leaf nodes, and depth ranges from 9 to 74402, 5 to 57919, and 4 to 19, respectively. See Table \ref{tab:all_hierarchies} for a more comprehensive overview of the tree properties. The hierarchies ImageNet-1K and Pizza are the exemplar ontologies employed in \cite{ayoughi2025designing}, and the other 14 hierarchies are the trees in the HierVision \cite{kasarlahiervision} repository that have at least a depth of 4, which gives room for restructuring. These hierarchies come from diverse domains such as biology, general object recognition, robotics, advertising, material science, actions/video recognition, and scenes/places recognition, and are visualized in Figure \ref{fig:trees}. 
\paragraph{\textbf{Hyperparameters}}
We use the GPT-4o model from OpenAI and DeepSeek-V3 as the Large Language Models in our experiments. In the hyperbolic embedding of hierarchies, each embedding algorithm is configured using its recommended hyperparameter settings from the corresponding papers. For the construction-based approaches, following Sala et al. \cite{sala2018representation}, the scaling factor $\tau$ is set to \(\tau = \frac{1}{1.3 * \ell} \log\Big(\frac{2 - \frac{\epsilon}{2}}{\frac{\epsilon}{2}}\Big)\), where $\epsilon$ is the machine precision of the applied floating point format and $\ell$ is the maximum path length of the tree, to avoid numerical problems while still obtaining near-optimal results. For HS-DTE, we adopt the settings from the original paper \cite{van2025low}. We train using projected gradient descent for $450$ iterations with a learning rate of $0.01$, reduced by a factor of $10$ every $150$ steps.
The minimum embedding size of each hierarchy is determined by the maximum degree of the tree in the Hadamard method, according to this equation: \(2^{\lfloor \log_2 n \rfloor} \geq \deg_{\max}(V)\). Using this equation, we determine the embedding size by further rounding the number to the closest multiple of 10. Table \ref{tab:all_hierarchies} shows the embedding size of all hierarchies and their corresponding maximum degree.

\paragraph{\textbf{Hyperbolic embedding evaluation metrics}}
Following the conventions in hyperbolic embedding literature \cite{nickel2017poincare,sala2018representation,sarkar2011low, ayoughi2025designing}, we focus on two metrics to evaluate the quality of tree embeddings. The first metric is average relative distortion, which measures the average relative embedding error between all pairs of nodes in $V$, given as follows for $N = |V|$ nodes:
\begin{equation}
    D_{avg} (\phi) = \frac{1}{N(N-1)} \sum_{u \neq v} \frac{|d_{\mathbb{D}} (\phi(u), \phi(v)) - d_T (u, v)|}{d_T (u, v)}.
\end{equation}
This metric measures how much the hyperbolic distance in the embeddings differs from the tree distance between all node pairs. The second metric is worst-case distortion, which specifically measures the ratio between the largest stretching and shrinking factor of pairwise distances:
\begin{equation}
    D_{wc} (\phi) = \max_{u \neq v} \frac{d_\mathbb{D}(\phi(u), \phi(v))}{d_T (u, v)} \bigg( \min_{u \neq v} \frac{d_\mathbb{D}(\phi(u), \phi(v))}{d_T (u, v)} \bigg)^{-1}.
\end{equation}
Where the average distortion measures the global distortion, the worst-case distortion captures large local distortions. A third commonly used metric is mean average precision (MAP), which measures how well local neighborhoods are preserved in the embedding space. However, since construction-based methods consistently achieve the best possible MAP value of $1.00$, we do not report it here. Due to the scale of some hierarchies, such as ImageNet-21K and Visual Genome, we re-implemented the distortion function to calculate distortion in batches and avoid out-of-memory errors. 

\begin{table*}[t]
\centering
\begingroup 
\setlength{\tabcolsep}{2.5pt}
\caption{\textbf{Hyperbolic embedding quality (ChatGPT | DeepSeek) and tree properties of all hierarchies. The numbers are bolded when restructuring performs better than the original hierarchy.}}
\label{tab:all_hierarchies}
\resizebox{1\linewidth}{!}{
\begin{tabular}{lcccccccccccc}
\toprule
 & \multicolumn{2}{c}{\textbf{Hadamard}} & \multicolumn{2}{c}{\textbf{HS-DTE}} & \multicolumn{8}{c}{\textbf{Tree Properties}}\\
\cmidrule(lr){2-3} \cmidrule(lr){4-5} \cmidrule(lr){6-13}
 & \multicolumn{1}{c}{$D_{avg}$} & \multicolumn{1}{c}{$D_{wc}$} & \multicolumn{1}{c}{$D_{avg}$} & \multicolumn{1}{c}{$D_{wc}$}& \multicolumn{1}{c}{Domain} & \multicolumn{1}{c}{$|V(G)|$} & \multicolumn{1}{c}{$|E(G)|$} & \multicolumn{1}{c}{Depth} & \multicolumn{1}{c}{\#Leaves} & \multicolumn{1}{c}{$\Delta(G)$} & \multicolumn{1}{c}{$d$} & \multicolumn{1}{c}{Avg BF} \\ \midrule
 ImageNet-1K\cite{deng2009imagenet} & 0.297 & 1.647 & 0.183 & 1.497 & General & 1778 & 1777 & 13 & 1015 & 26 & 40 & 2.3 \\
 \rowcolor{Gray} LLM-guided restructuring & \textbf{0.270} | \textbf{0.220} & \textbf{1.485} | \textbf{1.344} & \textbf{0.175} |  \textbf{0.151}& \textbf{1.434} | \textbf{1.329} &  & 1687 & 1686 & 12 & 1015 & 26 & 40 & 2.5 \\
Pizza\cite{ayoughi2025designing} & 0.126 & 1.180 & 0.087 & 1.144 & Robotics & 100 & 99 & 7 & 78 & 23 & 70 & 4.5 \\
 \rowcolor{Gray} LLM-guided restructuring & \textbf{0.065} | \textbf{0.089} & \textbf{1.090} | \textbf{1.118} & \textbf{0.051} | \textbf{0.060} & \textbf{1.091} | \textbf{1.103} &  & 89 & 88 & 4 & 81 & 39 & 70 & 11.0 \\
Core50 unbalanced\cite{fellbaum2010wordnet} & 0.089 & 1.112 & 0.057 & 1.072 & Robotics & 66 & 65 & 5 & 50 & 6 & 70 & 4.0 \\
\rowcolor{Gray}LLM-guided restructuring & \textbf{0.039} | \textbf{0.037} & \textbf{1.061} | \textbf{1.042} & \textbf{0.033} | \textbf{0.029} & \textbf{1.059} | \textbf{1.042} &  & 54 & 53 & 3 & 50 & 46 & 70 & 13.2\\
Core50\cite{lomonaco2017core50} & 0.075 & 1.092 & 0.037 & 1.048 & Robotics & 70 & 69 & 4 & 50 & 6 & 10 & 3.4\\
\rowcolor{Gray}LLM-guided restructuring & \textbf{0.052} | \textbf{0.053} & \textbf{1.064} | \textbf{1.064} & \textbf{0.034} | \textbf{0.033} & \textbf{1.043} | \textbf{1.046} &  & 63 & 62 & 3 & 50 & 6 & 10 & 4.7\\
MAdVerse\cite{sagar2024madverse} & 0.078 & 1.090 & 0.065 & 1.096 & Advertising & 656 & 655 & 4 & 578 & 26 & 130 & 8.3\\
\rowcolor{Gray}LLM-guided restructuring & \textbf{0.058} | \textbf{0.058} & \textbf{1.064} | \textbf{1.064} & \textbf{0.049} | \textbf{0.049} & \textbf{1.072} | \textbf{1.077} &  & 642 & 641 & 3 & 579 & 80 & 130 & 10.1\\
Matador\cite{beveridge2025hierarchical} & 0.094 & 1.119 & 0.053 & 1.070 & Materials & 82 & 81 & 5 & 59 & 9 & 40 & 3.5\\
\rowcolor{Gray}LLM-guided restructuring & \textbf{0.048} | \textbf{0.070} & \textbf{1.057} | \textbf{1.087 }& \textbf{0.042} | \textbf{0.051} & \textbf{1.061} | 1.076 &  & 66 & 65 & 3 & 59 & 20 & 40 & 9.2\\
Moments in Time\cite{monfort2019moments} & 0.074 & 1.092 & 0.057 & 1.108 & Video & 486 & 485 & 4 & 339 & 44 & 70 & 3.2\\
\rowcolor{Gray}LLM-guided restructuring & \textbf{0.069} | \textbf{0.054} & \textbf{1.087} | \textbf{1.064} & 0.065 | \textbf{0.045} & 1.124 | \textbf{1.082} &  & 486 & 485 & 4 & 438 & 53 & 70 & 10.1\\
BioTrove-LifeStages\cite{yang2024biotrove} & 0.096 & 1.148 & 0.024 & 1.051 & Biology & 19 & 18 & 6 & 5 & 4 & 10 & 1.2\\
\rowcolor{Gray}LLM-guided restructuring & \textbf{0.090} | \textbf{0.049} & \textbf{1.144} | \textbf{1.078} & 0.027 | \textbf{0.021} & 1.054 | \textbf{1.035} &  & 16 & 15 & 6 & 5 & 4 & 10 & 1.3\\
Marine Tree\cite{boone2022marine,boone2022mask} & 0.088 & 1.118 & 0.063 & 1.104 & Biology & 79 & 78 & 5 & 62 & 16 & 40 & 4.5\\
\rowcolor{Gray}LLM-guided restructuring & \textbf{0.049} | \textbf{0.066} & \textbf{1.064} | \textbf{1.087} & \textbf{0.038} | \textbf{0.053} & \textbf{1.071} | \textbf{1.092} &  & 75 & 74 & 3 & 62 & 22 & 40 & 5.6\\
NABirds\cite{van2015building} & 0.077 & 1.090 & 0.061 & 1.122 & Biology & 1011 & 1010 & 4 & 555 & 40 & 70 & 2.2\\
\rowcolor{Gray}LLM-guided restructuring & \textbf{0.057} | \textbf{0.057} & \textbf{1.064} | \textbf{1.064} & \textbf{0.043} | \textbf{0.043} & \textbf{1.087} | \textbf{1.087} &  & 999 & 998 & 3 & 555 & 50 & 70 & 2.2\\
COCO-10K\cite{caesar2018coco} & 0.166 & 1.232 & 0.093 & 1.139 & General & 234 & 233 & 8 & 171 & 10 & 20 & 3.6\\
\rowcolor{Gray}LLM-guided restructuring & \textbf{0.101} | \textbf{0.101} & \textbf{1.122} | \textbf{1.122} & \textbf{0.059} | \textbf{0.059} & \textbf{1.082} | \textbf{1.083} &  & 227 & 226 & 5 & 173 & 12 & 20 & 4.1\\
EgoObjects\cite{zhu2023egoobjects,ayoughi2025continual} & 0.321 & 1.905 & 0.170 & 1.594 & Robotics & 1665 & 1664 & 14 & 1179 & 22 & 40 & 3.4\\
\rowcolor{Gray}LLM-guided restructuring & \textbf{0.261} | \textbf{0.243} & \textbf{1.490} | \textbf{1.432} & 0.184 | \textbf{0.147} & \textbf{1.444} | \textbf{1.350} &  & 1482 & 1481 & 12 & 1192 & 22 & 40 & 5.1\\
OpenLoris\cite{she2020openloris,ayoughi2025continual} & 0.133 & 1.183 & 0.081 & 1.120 & Robotics & 98 & 97 & 7 & 69 & 10 & 20 & 3.3\\
\rowcolor{Gray}LLM-guided restructuring & \textbf{0.091} | \textbf{0.088} & \textbf{1.118} | \textbf{1.112} & \textbf{0.062} | \textbf{0.052} & \textbf{1.090} | \textbf{1.092} & &  95 & 94 & 5 & 69 & 10 & 20 & 3.6\\
PascalVOC\cite{everingham2010pascal} & 0.107 & 1.155 & 0.050 & 1.072 & General & 36 & 35 & 6 & 21 & 5 & 20 & 2.3\\
\rowcolor{Gray}LLM-guided restructuring & \textbf{0.047} | \textbf{0.088} & \textbf{1.064} | \textbf{1.122} & \textbf{0.030} | \textbf{0.044} & \textbf{1.042} | \textbf{1.060} &  & 27 & 26 & 3 & 21 & 8 & 20 & 4.3\\
ImageNet-21K\cite{ridnik2021imagenet} & 0.119 & 8.770 & 0.050 & 3.863 & General & 74402 & 74401 & 19 & 57919 & 401 & 520 & 4.5\\
\rowcolor{Gray}LLM-guided restructuring & 0.159 | 0.394 & \textbf{2.254} | 544.817 & 0.262 | 0.485 & \textbf{2.549} | 605.171 &  & 68205 & 68204 & 16 & 57919 & 401 & 520 & 6.6\\
Visual Genome\cite{krishna2017visual} & 0.132 & 392.641 & 0.174 & 2.930 & Scenes & 10503 & 10502 & 18 & 6114 & 107 & 130 & 2.3\\
\rowcolor{Gray}LLM-guided restructuring & 0.237 | 0.468 & \textbf{2.264} | \textbf{366.029} & 0.446 | 0.696 & \textbf{1.919} | 506.909 & & 8018 & 8017 & 15 & 6114 & 107 & 130 & 4.2\\
\bottomrule
\end{tabular}
}%
\endgroup
\end{table*}

\section{Experiments}
\subsection{Comparison}

In this experiment, we evaluate the effect of LLM-guided hierarchy restructuring on embedding quality. Table~\ref{tab:all_hierarchies} reports the average and worst-case distortion for two state-of-the-art construction-based methods across all hierarchies, along with key tree properties. Columns $|V(G)|$, $|E(G)|$, $\Delta(G)$, $d$, and \textit{Avg BF} refer to the number of nodes, number of edges, maximum degree of the graph, embedding size, and average branching factor, respectively. The even rows report the hyperbolic embedding quality of LLM-guided restructuring with ChatGPT and DeepSeek (ChatGPT | DeepSeek). To our knowledge, this is the first large-scale evaluation of SOTA hyperbolic embeddings—particularly on datasets such as ImageNet-21K and Visual Genome, which contain approximately $74.5\text{K}$ and $10.5\text{K}$ nodes, with minimum depths of $18$.

As shown in the table, LLM-guided restructuring with DeepSeek almost consistently outperforms the original hierarchy embedding quality, with the exception of the Matador, ImageNet-21K, and Visual Genome hierarchies. We observe that the OpenAI model improves the worst-case distortion for all embedding methods, except in BioTrove-LifeStages and Moments in Time hierarchies using HS-DTE. It also reduces the average distortion in most cases, with the exception of the Visual Genome and ImageNet-21K hierarchies under the Hadamard method, as well as the EgoObjects, BioTrove-LifeStages, and Moments in Time hierarchies under HS-DTE. Overall, the HS-DTE method tends to yield better embedding quality, while the Hadamard method offers faster runtimes.

\begin{figure*}[h]
  \centering
  \includegraphics[width=0.49\textwidth, trim={0cm 0cm 0cm 0cm}, clip]{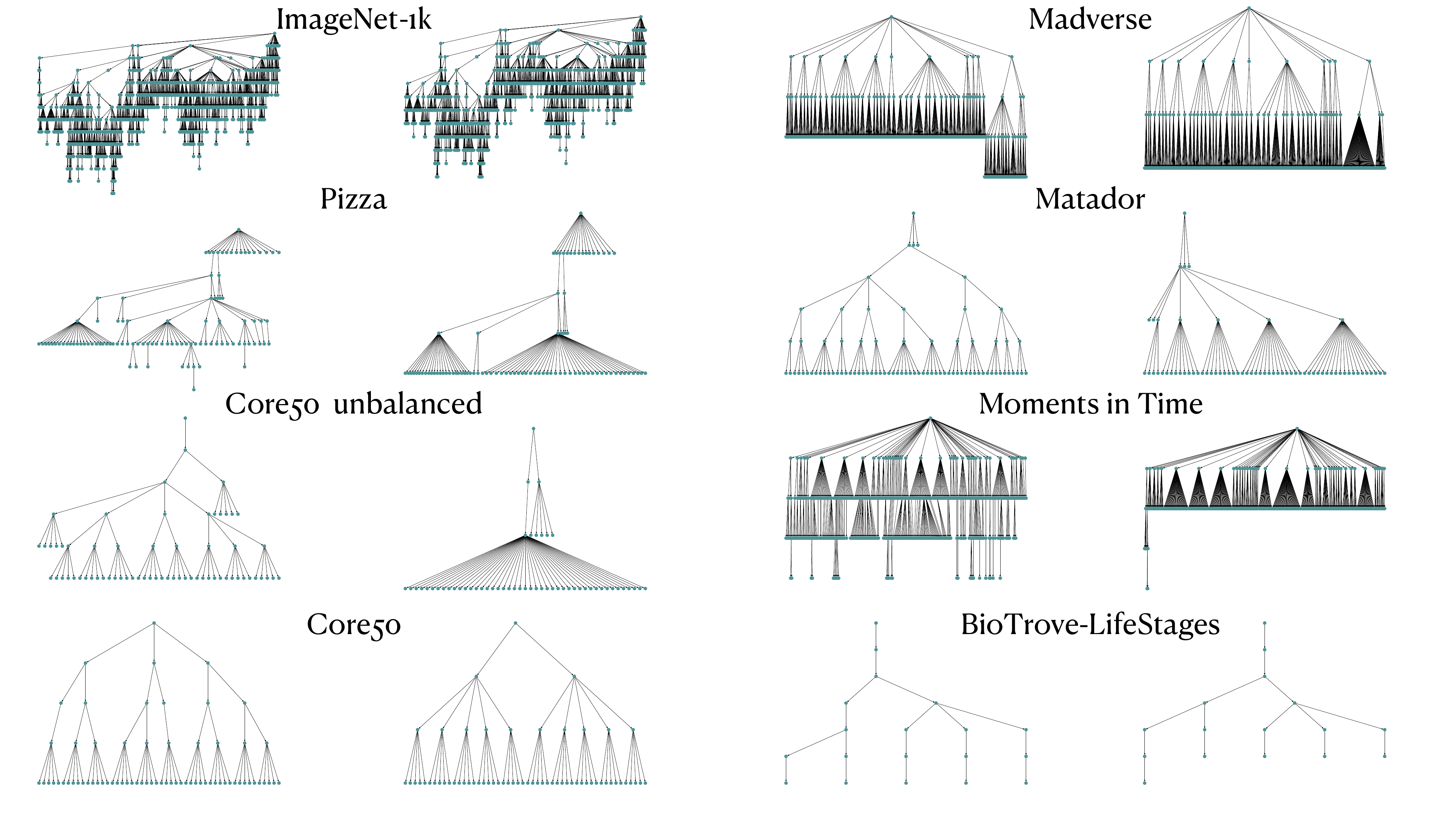}
\includegraphics[width=0.49\textwidth, trim={0cm 0cm 0cm 0cm}, clip]{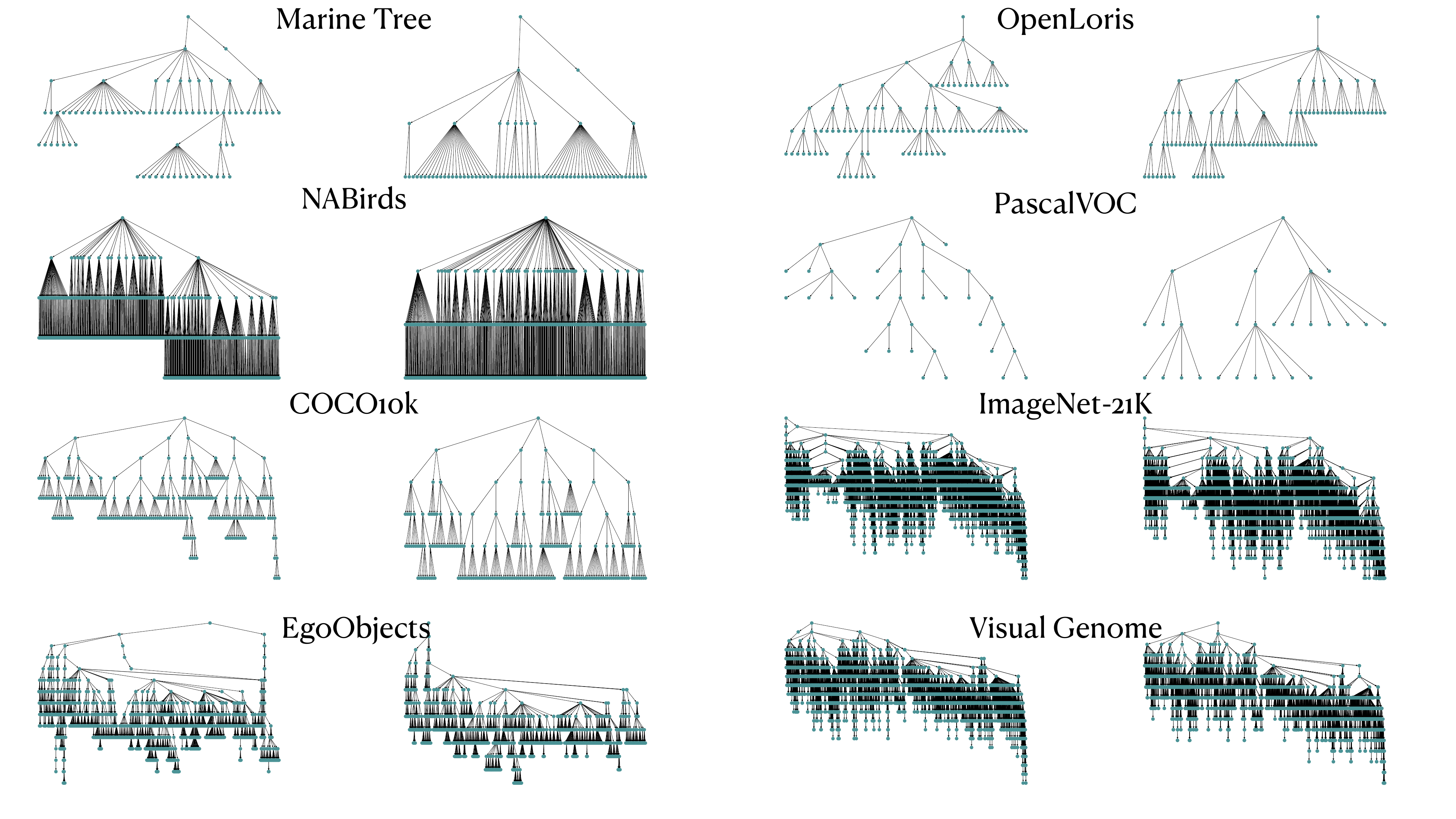}
  \caption{\textbf{Hierarchies before (left) and after (right) the LLM-guided restructuring}}
  \label{fig:trees}
\end{figure*}

When comparing tree properties, we observe that LLM-guided restructuring generally reduced the number of nodes across hierarchies. A notable exception is the Moments in Time hierarchy, where nodes were only promoted to higher levels, and no intermediate nodes were removed. The number of leaf nodes either remained the same—due to the removal of intermediate nodes—or increased when promoted leaves became siblings of intermediate nodes, which then became new leaf nodes. The maximum degree either stayed the same or increased, the average branching factor consistently increased, and the depth either remained unchanged or decreased. These changes indicate that the restructuring favors higher branching factors by flattening the tree structure and eliminating intermediate nodes, and promoting nodes. We find a correlation of -0.15 between the difference in branching factor and embedding distortion; higher branching factors lead to lower distortions. These changes are visible per hierarchy in Figure \ref{fig:trees}.



\subsection{Ablation study on prompt components}
The improved hyperbolic hierarchical embeddings are a result of four recommendations. To evaluate the effect of the recommendations in isolation and in combination, we have performed an ablation study on two hierarchies: COCO-10K and the Pizza hierarchy. The results are shown in Table \ref{tab:ablation_prompt}. We find that all recommendations individually improve distortion. Naturally, the first recommendation obtains the best improvements as it refers to a core principle of hyperbolic embeddings already observed in Table \ref{tab:all_hierarchies}: width over depth. The recommendations are also complementary, with the best results on average obtained when using all recommendations. We conclude that all recommendations matter for minimizing embedding distortion.

\begin{table}[t]
\centering
\begingroup 
\small
\renewcommand{\arraystretch}{0.9}
\setlength{\tabcolsep}{2.5pt}
\caption{\textbf{Which prompt combination is most important?}}
\label{tab:ablation_prompt}
\resizebox{1\linewidth}{!}{
\begin{tabular}{cccccccccccc}
\toprule
\multicolumn{4}{c}{} & 
\multicolumn{4}{c}{\textbf{COCO-10K}} & 
\multicolumn{4}{c}{\textbf{Pizza}} \\
\cmidrule(lr){5-8} \cmidrule(lr){9-12}
\multicolumn{4}{c}{\small\textbf{Recommendation}} & 
\multicolumn{2}{c}{\small\textbf{Hadamard}} & 
\multicolumn{2}{c}{\small\textbf{HS-DTE}} & 
\multicolumn{2}{c}{\small\textbf{Hadamard}} & 
\multicolumn{2}{c}{\small\textbf{HS-DTE}} \\
\cmidrule(lr){1-4} \cmidrule(lr){5-6} \cmidrule(lr){7-8} \cmidrule(lr){9-10} \cmidrule(lr){11-12}
$R_1$ & $R_2$ & $R_3$ & $R_4$ & $D_{avg}$ & $D_{wc}$ & $D_{avg}$ & $D_{wc}$ & $D_{avg}$ & $D_{wc}$ & $D_{avg}$ & $D_{wc}$ \\ 
\midrule
&&&&0.166 & 1.232 & 0.093 & 1.139& 0.126 & 1.180 & 0.087 & 1.144\\
\checkmark &        &        &        & 0.139& 1.188  &    0.090    &   1.144     &        0.069 &     1.090&     0.055   &    1.089  \\
          & \checkmark &        &        &      0.143    &    1.189   &   0.084     &   1.128     &   0.034     &     1.042   &  0.031      &   1.054     \\
          &        & \checkmark &        &         0.169    &   1.232&    0.087    &  1.125      &        0.128&   1.184     &    0.085    &   1.139     \\
\checkmark & \checkmark & \checkmark &        &      0.101   &   1.122&    0.059    & 1.086&   0.105     &    1.148    &    0.076    &    1.134   \\
          & \checkmark & \checkmark & \checkmark &     0.076    &    1.092  &  0.055      &    1.078    &  0.127   &    1.180    &     0.086   &     1.139   \\
\checkmark & \checkmark & \checkmark & \checkmark &0.101 & 1.122 & 0.059 & 1.082 & 0.065 & 1.090 & 0.051 & 1.091\\
\bottomrule
\end{tabular}
}
\endgroup
\end{table}

\subsection{Explainability}
\label{sec:explainability}

An important aspect of this approach is that SOTA LLMs can produce explanations of the steps they are taking in an interactive fashion. Here, we demonstrate how this is used. After being given the initial prompt, the LLM responds by dividing the task into smaller subtasks. First, it starts by providing an assessment of the current hierarchy structure. For the Core50 hierarchy, the LLM provided the following:

\begin{chatgptbox}
The hierarchy is structured as a deep tree:\\
* Root node: 50\\
* Intermediate nodes: 51, 52, 54, etc.\\
* Leaf nodes (classes): 0–49\\
* Depth: $\sim$6 levels\\
* Each leaf is reached via a long path through multiple intermediate nodes.
\end{chatgptbox}
Subsequently, the LLM goes through each recommendation in the prompt and checks if it applies to the hierarchy and makes an action plan accordingly. The following is an example from the ImageNet-1K hierarchy. 

\begin{chatgptbox}\textbf{Recommendation 1:} Design for Width, Not Depth\\
“Algorithms perform best with wide hierarchies that have high branching factors.”
Observation: Your hierarchy has many deep, narrow chains like:
1777 → 693 → 10 → 2 → 1 → 0 and
210 → 209 → 182 → 181 → 177 → 173 → 172
Improvement Suggestions:\\
* Flatten deep chains where intermediate nodes have a single child and no additional semantic weight.
For example, replace them with: 1777 → 0 [flattened from 693 → 10 → 2 → 1 → 0]\\
* Collapse nodes with single children and promote their descendants higher up, e.g.:
Before: 10 → 2 → 1 → 0 After: 10 → 2, 1, 0 or 10 → 2 → 1, 0\\
* Increase Branching Factor Near the Root
Your top-level node 1777 has relatively few direct children compared to what could be beneficial for embedding performance.
Before: 1777 → 693 → 10 → 2 After: 1777 → 693, 10, 2 \\
* If your ontology permits and semantics aren’t lost: (i) Linear chains of depth > 3 with no branching should be flattened. (ii) Try to favor a bushier structure, especially in early levels.\\

\textbf{Recommendation 2:} Ignore Tree Balance\\
“Embedding performance is not significantly impacted by imbalance.”
Observation: Your hierarchy is naturally imbalanced — some branches go 10+ levels deep, while others are only 2–3.
No Action Needed: You can retain imbalance where semantically meaningful.\\

\textbf{Recommendation 3:} Node Complexity is Acceptable\\
“More nodes only moderately impact distortion.”
Observation: Your hierarchy includes hundreds of nested classes (e.g., under 574 → 573 → 210 etc.), many of which have several siblings.
Conclusion: Your node count is not a limitation. You can keep the current granularity if it's meaningful for your embeddings.\\

\textbf{Recommendation 4:} Avoid Multiple Inheritance\\
“Best to have single inheritance.”
Observation: From inspection, your hierarchy seems strictly tree-structured, with no node appearing in multiple branches (i.e., no multiple parents).
No Action Needed: Structure is ideal for hyperbolic embedding models.
\end{chatgptbox}
Another example from the Pizza hierarchy:
\begin{chatgptbox}
\textbf{R1. Design for Width, Not Depth}\\
* Flatten some of the deeper branches.
Before: VegetableTopping → PepperTopping  → GreenPepperTopping → HotGreenPepperTopping
After: VegetableTopping → PepperTopping → HotGreenPepperTopping, SweetPepperTopping, JalapenoPepperTopping \\
* Consider splitting PizzaTopping into more first-level categories if appropriate (e.g., Grouping by taste profiles, like "SpicyToppings", "SweetToppings", etc.).\\

\textbf{R4. Avoid Multiple Inheritance}\\
Observation: Some pizzas may conceptually belong to multiple categories (e.g., SpicyPizza, VegetarianPizza, RealItalianPizza, etc.). If any class instance (e.g., Capricciosa) is listed under more than one parent (multiple inheritance), it could increase embedding distortion.
Recommendations:
Try to use single inheritance and move the multi-class information to annotations or properties instead. E.g., instead of Capricciosa inheriting both from NamedPizza and SpicyPizza, let it only inherit from NamedPizza, and annotate it as hasSpiciness = Medium or isSpicy = true.
\end{chatgptbox}

When analyzing the suggestions produced by the LLM for all hierarchies, common suggestions include: chain removal, node removal, and node promotions. Finally, the LLM makes a summary of the proposed changes and gives options for the knowledge engineer to choose from:

\begin{chatgptbox}
Would you like me to:\\
* Output a flattened version of this hierarchy text with improved width?\\
* Generate a Python script to automate restructuring?\\
* Visualize the before-and-after hierarchy structure?\\
* Provide a fully revised version of the hierarchy based on these principles?\\
* Let me know how you'd like to proceed!\\
* Let me know if you'd like an even flatter structure or want to adjust category groupings.
\end{chatgptbox}

Once asked to apply the changes, the LLM will decide to apply the changes using two different approaches. The first approach is to directly predict the restructured hierarchy with next token generation and is usually used for smaller hierarchies such as PascalVOC and BioTrove-LifeStages. This approach could also be used for bigger hierarchies such as Moments in Time. However, due to LLM length constraints, the reply has been divided into multiple chat outputs. The second approach involves predicting a function first, and then applying the function to the hierarchy, which has been applied to bigger hierarchies such as ImageNet-21K and Visual Genome. 
\vspace*{-3mm}
\section{Conclusion}
In this work, we developed an approach that leverages LLMs to help guide knowledge engineers in restructuring ontologies for improved performance in hyperbolic learning. Our approach leads to improved embedding quality across 16 different hierarchies. Crucially, the approach provides explanations to the knowledge engineer, guiding them through suggested changes, helping them make informed decisions about whether to apply them.

Future work includes investigating the generality of the approach for use with other large language models and incorporating the approach into knowledge engineering methodologies. Lastly, our goal is to not only assess the restructuring of existing ontologies but also the creation of new ones. We hope that this and other LLM-guided approaches will help knowledge engineers better incorporate the needs of downstream hierarchical learning tasks into the knowledge artifacts they create.


\begin{acks}
This work was partially supported by the EU's Horizon Europe research and innovation programme within the ENEXA project (grant Agreement no. 101070305).
\end{acks}

\bibliographystyle{ACM-Reference-Format}
\bibliography{sample-base}


\end{document}